\documentclass[runningheads]{llncs}
\usepackage[T1]{fontenc}
\usepackage{graphicx}
\usepackage[table,dvipsnames]{xcolor} 
\usepackage{color}
\definecolor{Fraunhofergreen}{RGB}{23,156,125}
\definecolor{Fraunhofersteelblue}{RGB}{0,91,127}
\definecolor{Fraunhofersilvergrey}{RGB}{166,187,200}
\definecolor{Fraunhoferorange}{RGB}{245,130,32}

\definecolor{Fraunhofergraphit}{RGB}{28,63,82} 
\definecolor{Fraunhofersand}{RGB}{211,199,174} 
\definecolor{Fraunhoferpetrol}{RGB}{0,133,152}
\definecolor{Fraunhoferaqua}{RGB}{57,193,205}
\definecolor{Fraunhoferlime}{RGB}{178,210,53}

\definecolor{Fraunhoferblue}{RGB}{31,130,192} 
\definecolor{Fraunhoferred}{RGB}{187,0,86} 
\definecolor{Fraunhoferweinrot}{RGB}{124,21,77} 

\definecolor{Fraunhofersteelblue}{RGB}{0,91,127}
\definecolor{Fraunhofersilvergrey}{RGB}{166,187,200}  
\definecolor{KITred}{RGB}{160,30,40}
\definecolor{KITblue}  {RGB}{ 70,100,170} 
\definecolor{KITblue70}{RGB}{125,146,195} 
\definecolor{KITblue50}{RGB}{162,177,212} 
\definecolor{KITblue30}{RGB}{199,208,229} 
\definecolor{KITblue15}{RGB}{227,232,242} 
\usepackage{pifont}

\newcommand{\cmark}{\ding{51}}%
\newcommand{\xmark}{\ding{55}}%
\newcommand{\cbox}[1]{\tikz[baseline=-0.5ex]\draw[#1, line width=3, ](0,0) -- (0.2, 0);}

\usepackage[]{todonotes}
\usepackage[backref=page,pagebackref=true,breaklinks=true,colorlinks,bookmarks=false, linkcolor=black, citecolor=Fraunhofergreen, urlcolor  = black]{hyperref}

\usepackage{tikz}
\usetikzlibrary{%
	automata,%
	arrows,%
	backgrounds,%
	calc,%
	chains,%
	decorations,%
	decorations.markings,%
	decorations.pathmorphing,%
	external,%
	fadings,%
	fit,%
	matrix,%
	positioning,%
	scopes,%
	shadows,%
	shapes.geometric,%
	shapes.misc,%
	shapes.multipart,%
	through,%
	mindmap, 
	backgrounds, 
	topaths}

\usetikzlibrary{arrows.meta}
\usetikzlibrary{decorations.pathreplacing}
\tikzset{
	pics/square/.default={1},
	pics/square/.style = {
		code = {
			\draw[pic actions, ultra thick] (0,0) rectangle (#1,#1);
		}
	}
}
\usepackage{multirow}

\usepackage{apalike}

\usepackage[square,sort,comma,numbers]{natbib}
\usepackage{makecell} 
\usepackage[bottom]{footmisc}
\usepackage{amsmath}
\usepackage{ifthen}
\newboolean{submission} 
\setboolean{submission}{false} 

\begin{document}

%
%
%
%
%
%
	
%
\title{Utilizing dataset affinity prediction in object detection to assess training data}
\titlerunning{Utilizing dataset affinity prediction in object detection}

\ifthenelse{\boolean{submission}}{          
\author{Anonymous submission}
\authorrunning{}
\institute{}
}
{
\author{Stefan Becker \inst{1}\href{https://orcid.org/0000-0001-7367-2519}{\includegraphics[scale=0.04]{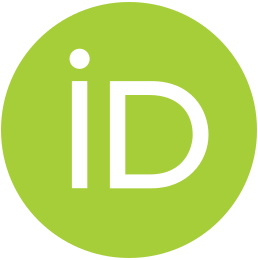}}\href{mailto:stefan.becker@iosb.fraunhofer.de}{\includegraphics[scale=0.04]{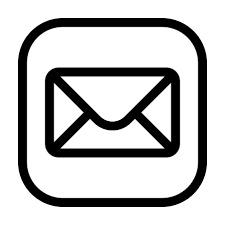}} 
	 \and
	Jens Bayer \inst{1}\href{https://orcid.org/0000-0002-2806-6920}{\includegraphics[scale=0.04]{img/orcid-og-image.jpg}}
	\and
	Ronny Hug \inst{1}\href{https://orcid.org/0000-0001-6104-710X}{\includegraphics[scale=0.04]{img/orcid-og-image.jpg}} \and 
	Wolfgang Huebner \inst{1}\href{https://orcid.org/0000-0001-5634-6324}{\includegraphics[scale=0.04]{img/orcid-og-image.jpg}}   \and \\ Michael Arens \inst{1}\href{https://orcid.org/0000-0002-7857-0332}{\includegraphics[scale=0.04]{img/orcid-og-image.jpg}}
}
\authorrunning{S. Becker et al.}
\institute{Fraunhofer IOSB, Ettlingen, Germany\\
	Fraunhofer Center for Machine Learning \\
	\email{\{firstname.lastname\}@iosb.fraunhofer.de}\\
	\url{www.iosb.fraunhofer.de}}
} 
\maketitle              
\begin{abstract}
Data pooling offers various advantages, such as increasing the sample size, improving generalization, reducing sampling bias, and addressing data sparsity and quality, but it is not straightforward and may even be counterproductive. Assessing the effectiveness of pooling datasets in a principled manner is challenging due to the difficulty in estimating the overall information content of individual datasets. Towards this end, we propose incorporating a data source prediction module into standard object detection pipelines. The module runs with minimal overhead during inference time, providing additional information about the data source assigned to individual detections. We show the benefits of the so-called dataset affinity score by automatically selecting samples from a heterogeneous pool of vehicle datasets. The results show that object detectors can be trained on a significantly sparser set of training samples without losing detection accuracy. 

\end{abstract}
%
%
%
%
\section{Introduction and Related Work}
\label{sec:introduction}
Despite their growing scale, single datasets capture only limited visual aspects of a target domain, and obtaining more label data takes time and effort. One way to overcome these limitations is data pooling, the combination of datasets. Combining datasets  results in a larger sample size, providing more instances for model training, which can lead to more robust and generalizable models, especially in situations where the original datasets are relatively small. Data pooling also aims to increase diversity in the data to avoid overfitting to specific patterns present in one dataset but not in another, but also increase intra-class variations and scene variation. It can also be useful when dealing with imbalanced datasets by combining datasets with different class distributions. However, an arbitrary combination of datasets capturing a specific object category does not guarantee an improved detection performance and can be counterproductive. For example, extending with samples too far away from the target domain can lead to a decline in the detector's performance. Combining strongly correlated datasets can be redundant and provides no new information.

One central problem is that the effectiveness of pooling datasets can not be assessed in a principled way because the overall information content of individual datasets is hard to estimate. Towards this end, we propose extending standard object detection pipelines with an additional inference head to predict a dataset affinity score. The affinity prediction assigns every detection to the set of pooled datasets realized as a multinomial logistic regression task. Thus scoring the affinity between detections and training datasets. This enables direct model-depend feedback to the training data during inference. Thereby, we gain information on which dataset contributed to individual detections. This can be seen as a kind of ante-hoc detection explanation. We use the affinity scores to identify datasets that support the overall detection performance and datasets where the scores suggest a higher domain gap between the training and target sets. Based on the assigned dataset affinity distribution, we prune the training set and show that detectors trained on a significantly sparser set achieve similar detection accuracy. By providing the dataset affinity score during inference, our approach stands in contrast to post-hoc explanation methods, producing visual explanations. 

These methods specify saliency maps to interpret the object predictions. There are gradient-based methods such as Grad-CAM \cite{Selvaraju_ICCV_2017} or GradCAM++ \cite{Chattopadhay_WACV_2024} where the saliency maps are based on the gradient of the model's output with respect to the input features and perturbation-based methods where generating these maps involve perturbing or altering input features and observing the impact on the model's prediction (e.g., RISE \cite{Petsiuk_BMVC_2018}, D-RISE \cite{Petsiuk_CVPR_2020}, LIME \cite{Ribeiro_SIGKDD_2016}). For a more detailed view of explainable artificial intelligence (XAI), including post-hoc explanation methods, we refer to the survey of Burkart et al. \cite{burkart2021survey} and for a comparison of several saliency maps generation methods for object detection to Bayer et al. \cite{Bayer_SPIE_2022}. 

The dataset affinity score provides information on whether an object in the target domain can be explained with samples from particular datasets. So, assessing the model performance during processing with estimating the detector's uncertainty can be seen as an alternative concept of getting direct feedback on the target domain. Several techniques have been proposed to integrate uncertainty estimation into a detector. There are approaches using \emph{variational inference} by relying on \emph{Bayesian neural networks} (BNNs) \cite{MacKay_NC_1992} or using \emph{Monte-Carlo dropout} \cite{Gal_ICML_2016} as more a practical way to perform approximate inference, such as the work of Azevedo et al. \cite{Azevedo_ML4AD_2020}. Then there are approaches using \emph{direct modeling} by assuming a certain probability distribution over the detector outputs (e.g., \cite{Choi_ICCV_2019}). Further, there are approaches estimating predictive probability using an ensemble of models where the outputs from each detector are treated as independent samples from a mixture model \cite{Lakshminarayanan_NeurIPS_2017}. For a more detailed view of uncertainty estimation in the context of object detection, we refer to the surveys of Feng et al.\cite{Feng_TITS_2020} and Hall et al.\cite{Hall_WACV_2020}. 

Other concepts to assess the efficiency of data pooling are not part of the model itself. These concepts use statistical measures like the \emph{Kullback-Leibler} divergence or \emph{Wasserstein} distance to quantify the difference between probability distributions of domains \cite{Hinton_Roweis_2003} or make a feature space analysis by examining the distribution of features in different domains with histograms, scatter plots, or kernel density estimation \cite{Goodfellow_book_2016}. Thus, these concepts use external post-processing steps to compare learned representation. This 
comparing and assessing dataset domain gaps, and therefor data pooling, is closely related to or rather part of the broader problem of dataset shift. 

Dataset shift is a concept that encompasses various distribution changes that can occur within or between domains, leading to the failure of even high-capacity models. Reasons for domain shift include seasonal or weather change. For a detailed overview of dataset shift and related sub-problems, we refer to the work of \cite{Quionero-Candela2009}. Domain adaptation is a specific technique to address dataset shifts in cases where the change in data distribution is due to a shift between domains.

In the context of domain adaptation for object detection, corresponding methods try to align the source domain distribution to a particular target domain. Some approaches \cite{Chen_CVPR_2020,Chen_CVPR_2018} try to learn invariant features by feature alignment via adversarial training \cite{Ganin_JMLR_2016}. Other methods try to align object instances across domains utilizing category-level centroids \cite{Zhu_CVPR_2019} or attention maps \cite{Vs_CVPR_2021}. Domain generalization aims to generalize to domains unseen in training. For example, the approach form Vidit et al. \cite{Vidit_CVPR_2023} leverages a pre-trained vision model to develop a semantic augmentation strategy for altering image embeddings.

Besides the problems of data pooling, training from multiple datasets also faces the problem of varying label sets. To align multiple datasets, we unify the label sets by mapping sub-categories to a subsuming super-category or, rather, a more general category. In our case, we subsume different land-vehicle types such as 'car, 'van', 'truck' under the super-category 'vehicle'. In the context of unifying label sets,  Redmon et al. \cite{Redmon_CVPR_2017} introduced a hierarchical model of visual concepts (WordTree) to combine the labels of ImageNet \cite{Russakovsky_IJCV_2015} and MS COCO \cite{Lin_ECCV_2014}. ImageNet labels are pulled from WordNet \cite{Miller_ACM_1995}, a language database that structures concepts and their relation. Redmon et al. \cite{Redmon_CVPR_2017} utilizes several classification scores over co-hyponyms of the WordTree to realize a more fine-grained object classification along the hierarchical label tree. Nevertheless, considering a hierarchical tree with different levels of information and intra-class differences, we follow the concept of mapping all intra-class variations of vehicle classes to one comprehensive super-category. Merging datasets this way for an object category has already been proven to learn more general and robust models. For example, Hasan et al. \cite{Hasan_CVPR_2021} combined multiple pedestrian datasets, showing improved cross-dataset performance. For segmentation, Lampert et al. \cite{Lambert_CVPR_2020} merged and split different classes from datasets to realize a unified flat taxonomy to be still compatible with the standard training method. 

Although approaches that learn a label space from visual data go beyond this paper's scope, we also mention a few to cover this aspect. The task is considered universal representational or universal detectors. Another approach toward a universal detector is the work of Wang et al. \cite{Wang_CVPR_2019} They proposed to train a detector from multiple datasets in a multi-task setting. Zhao et al. \cite{Zhao_ECCV_2020} train a universal detector on multiple datasets by manually merging the taxonomies and train with cross-datasets pseudo-labels generated by dataset-specific models. In the work of Zhou et al. \cite{Zhou_CVPR_2022a} they fuse multiple annotated datasets without manually merging by formulating an optimization problem on which dataset-specific output should be merged.

Since the proposed additional affinity prediction relies on an object detector, we refer to the following works \cite{Zhao_TNNLS_2019,Zou_arXiv_2019,Xiao_MTAA_2020,Liu_IJCV_2020,Wu_NC_2020,Jiao_IEEEAccess_2019}  for an overview on current trends and state-of-the-art models for object detection.

The main contribution of this paper is to present a new idea to assess the effectiveness of data pooling. We propose to extend detection pipelines with an additional inference head to predict the affinity to pooled training data sets. With minimal overhead, the affinity scores allow direct feedback to training samples during run-time. The score provides information on which dataset is responsible for explaining individual detections and the selection of a sparser training set without performance decrease.  

The paper is structured as follows. The next section provides a description of the proposed additional dataset affinity prediction (section \ref{sec:doc}). In section \ref{sec:dataset_alignment}, the selected datasets for training and their alignment are described. The evaluation and achieved results are discussed in section \ref{sec:evaluation}. Finally, a conclusion is given in section \ref{sec:conclusion}.

\section{Dataset Affinity Prediction}
\label{sec:doc}
To better assess the efficiency of data pooling, we propose to use an additional inference head to estimate the affinity to datasets in the data pool for every detection. Since current object detectors are designed in a way that they internally use separate heads for different inference tasks, this concept is applicable to almost all current detection pipelines. Given an image $\textbf{I}_{k}$ with index $k$ applying a modified detector with the additional dataset affinity score results in the following output: 
\begin{equation}	
	detector_{\Theta}(\textbf{I}_k)	\rightarrow  \{\vec{d}_{i,k}=(o, \vec{b}, \vec{c}, \vec{a}) \}^{N_{d,k}}_{i}
\end{equation}

$\Theta$ are the model parameters. The output is a set of $N_{d,k}$ detected objects $\vec{d}$ with object index $i$, where $o$ is the objectness or confidence score, $\vec{b}$ the object location description in the image (i.e., the bounding box with central point, width and height of the object $\vec{b}=\{b_x,b_y,b_w,b_h\}$), $\vec{c}$ the class labels, and $\vec{a}$ the dataset affinity scores where the dimension corresponds to number of datasets in the training pool. Adding the affinity score is stated as a multinomial logistic regression task to distinguish between the individual datasets of the combined training pool.

Here, we exemplarily build on a recent variant of the \emph{You Only Look Once} (YOLO) object detection family, in particular on the YOLOv7-X \cite{Wang_CVPR_2023} detector. YOLO is a so-called \emph{single shot detector}. This means that objects are detected in a single forward pass without additional steps such as \emph{region proposal networks} \cite{Girshick_ICCV_2015,Ren_TPAMI_2017}. Thus, YOLO variants are particularly suitable for real-time applications. YOLO variants use separate inference heads for localization and classification and thus fulfill the requirements to apply the proposed extension. During training, this is considered with multiple loss terms. In particular, YOLOv7-X uses an objectness loss $\mathcal{L}_{obj}$, a classification loss $\mathcal{L}_{cls}$, and a localization loss $\mathcal{L}_{loc}$ to form the complete loss function that guides the training process of the model. The objectness loss assists in accurate object localization and classification by distinguishing between cells that contain objects and those that do not. $\mathcal{L}_{loc}$ corresponds to the bounding box regression head that is responsible for refining the precise location and size of detected objects. The classification head and hence $\mathcal{L}_{cls}$ focuses on classifying detected objects into predefined categories. It typically involves using \emph{softmax} functions to assign each object to a specific class label. Relying on the same information as the classification head that distinguishes between object classes, a similar head is added that distinguishes between every dataset added in the training set. With this adaptation, the overall loss term, including the affinity loss $\mathcal{L}_{aff}$, is given by: 
\begin{equation}
	\mathcal{L} = \lambda_{obj}\mathcal{L}_{obj} + \lambda_{cls} \mathcal{L}_{cls} + \lambda_{loc} \mathcal{L}_{loc} + \lambda_{aff} \mathcal{L}_{aff} 
	\label{eq:loss} 
\end{equation}
Similar to Wang et al. \cite{Wang_CVPR_2023}, the weighting factors of the loss terms are set to $\lambda_{obj}=0.7$, $\lambda_{cls}=0.3$, $\lambda_{loc}=0.05$, and we set $\lambda_{aff}=0.3$ after a grid search. $\mathcal{L}_{objectness}$ uses \emph{binary cross entropy}. To calculate $\mathcal{L}_{loc}$ the \emph{complete intersection over union} (CIoU) is utilized. The classification loss and the affinity loss are realized using \emph{focal loss} \cite{Lin_ICCV_2017}. However, in the case of a single class detector such as for 'vehicle', also \emph{binary cross entropy} is used as classification loss. A schematic illustration of the proposed object detection pipeline is depicted in Figure \ref{fig:detection_pipeline}.

The model is implemented using \emph{Pytorch} \cite{Paszke_NEURIPS_2019} building on the YOLOv7-X detector implementation of \cite{Wang_CVPR_2023} \footnote{\url{https://github.com/WongKinYiu/yolov7} (accessed 14.11.2023)}. For training, an \emph{ADAM} optimizer variant \cite{Kingma_ICLR_2015,Loshchilov_ICLR_2019} with a starting learning rate of $0.001$ is used.
\input{figs/fig_detector_affinity} 
\section{Dataset Alignment}
\label{sec:dataset_alignment}
Besides getting insights into the training data pool, we follow the concept of combining datasets for an object category to learn more general and robust models. A problem that arises from this is differing label sets. Concepts of unifying label sets have already been discussed in section \ref{sec:introduction}. To build a general 'vehicle' detector, we map different sub-categories of vehicles to the more general parent class or rather super-category for aligning the datasets in terms of object labels. In addition to the class labels used, we categorize vehicle datasets according to two criteria: dataset types and sensor positions (viewing angle) during data recording. For other dataset characteristics, we refer to the following reviews and surveys \cite{Bogdoll_Vehits_2022,Bogdoll_arxiv_2023,Janai2020,Laflamme_arxiv_2019,Yin_ITSC_2018,Long_JSTEORS_2021,Song_arxiv_2023,Danaci_arxiv_2023,Yurtsever_IEEEAcces_2020}.

For dataset types, we distinguish between \emph{general} datasets and \emph{domain-specific} datasets. \emph{General} datasets, also called \emph{foundation data}, are designed to capture a diverse range of objects or scenes. Examples of such datasets include ImageNet \cite{Russakovsky_IJCV_2015}, MS COCO \cite{Lin_ECCV_2014}, and OpenImages \cite{openimages2020Kuznetsova}. \emph{General} datasets typically contain a large number of diverse images with a broad range of object categories, allowing researchers to test the performance of their models on a wide variety of objects and backgrounds. \emph{Domain-specific} or \emph{task-specific} datasets, on the other hand, are designed to capture a specific type of object or scene that is relevant to a particular domain or application. Examples of such datasets for the application domain of autonomous driving include FLIR \cite{flirDataset}, Cityscapes \cite{Cordts_CVPR_2016}, and KITTI \cite{Geiger_CVPR_2012}. These datasets are often smaller in size compared to general datasets, but they are curated to capture the specific challenges and characteristics of the domain or task, and have only a small set of class labels. The advantage of using \emph{domain-specific} datasets is that they are tailored to the specific requirements and constraints of the application or domain. However, this restriction may hinder an object detector trained on these datasets from generalizing to other domains.

An additional difference between \emph{general} and \emph{domain-specific} datasets is that \emph{general} datasets consist of randomly pooled image collections instead of data recorded with a specific sensor. For example, the image sensors for autonomous driving. Despite the extremely large variation these datasets have to capture, the sensor position is always close to ground-level with a specific viewing angle of the scene. Thus, our next criterion to categorize datasets is the sensor position corresponding to the sensor platform or the altitude of the sensor platform. These are ground-level datasets captured from car sensors or body cams. Then, there are low, mid, and high-altitude datasets. Low-altitude datasets are commonly captured with fixed surveillance cams and are widespread in the application domain of traffic monitoring. Mid-altitude datasets often come from the same application domain but are captured with small UAVs. Lastly, high-altitude datasets or aerial datasets where the data is recorded with a sensor on a satellite or high-flying drones, etc. 
\input{figs/fig_datasets} 
\begin{table*}[h!]
	\caption{Key characteristic of the aligned datasets used for training a general 'vehicle' detector.}
	\label{tab:selection_datasets} 
	\centering
	\resizebox{1.0\textwidth}{!}{
		\begin{tabular}{|c|c|c|c|c|c| >{\centering}m{6cm} |c|}
			\hline
			{\bf dataset} &dataset type & resolution / pixel &\# images & \# aligned images & \# categories &  \# vehicle categories & \# instances  \\
			\hline
			\rowcolor{gray!20}
			&   &  &  &   & & $4$  &   \\
		   	\rowcolor{gray!20}
		    \multirow{-2}{*}{MS COCO \cite{Lin_ECCV_2014} }  &  \multirow{-2}{*}{general} &  \multirow{-2}{*}{$640\times640$}  &  \multirow{-2}{*}{$328.0k$} &  \multirow{-2}{*}{$118.3k$} &  \multirow{-2}{*}{$80$} &  ('car', 'motorcycle', 'truck', 'bus')  &  \multirow{-2}{*}{$68634$} \\
		    	\rowcolor{gray!10}
		    &   &  &  &   & & $4$  &   \\
			\rowcolor{gray!10}    
			 \multirow{-2}{*}{DETRAC  \cite{Wen_CVIU_2020}}&  \multirow{-2}{*}{domain-specific} &  \multirow{-2}{*}{$960\times540$} &  \multirow{-2}{*}{$84.0k$} &  \multirow{-2}{*}{$8.1k$} &  \multirow{-2}{*}{$4$} &  ('car', 'van', 'bus', 'others') &  \multirow{-2}{*}{$46814$} \\
			 	\rowcolor{gray!20}
			 &   &  &  &  &  & $3$ &  \\
			\rowcolor{gray!20}
			\multirow{-2}{*}{UAVDT  \cite{Du_ECCV_2018}} & \multirow{-2}{*}{domain-specific}   & \multirow{-2}{*}{$1024\times540$} & \multirow{-2}{*}{$80.0k$} & \multirow{-2}{*}{$4.1k$} & \multirow{-2}{*}{$3$} & ('car', 'truck', 'bus') &  \multirow{-2}{*}{$33942$} \\		
				\rowcolor{gray!10}
			 &   &  &  &  &  & $8$ &  \\
			\rowcolor{gray!10}
			 \multirow{-2}{*}{VisDrone \cite{Zhu_TPAMI_2021}} &  \multirow{-2}{*}{domain-specific}     &  \multirow{-2}{*}{$960\times540$} & \multirow{-2}{*}{$10.2k$}  & \multirow{-2}{*}{$1.0k$} & \multirow{-2}{*}{10} & (car', 'van', 'truck', 'tricycle',  'awning-tricycle', 'bus', 'motor', 'other')    & \multirow{-2}{*}{124977} \\
			\rowcolor{gray!20}
			 &   &  &  &   & & $6$  &   \\
			\rowcolor{gray!20}
			 \multirow{-2}{*}{FLIR VIS  \cite{flirDataset} }&   \multirow{-2}{*}{domain-specific }   &   \multirow{-2}{*}{$1800\times1600$ }&   \multirow{-2}{*}{$10.3k$ } &  \multirow{-2}{*}{$9.3k$ } &  \multirow{-2}{*}{15 } & ('car', 'motor', 'bus', 'truck',  'scooter', 'other vehicle')   &  \multirow{-2}{*}{76946} \\
			\rowcolor{gray!10}
			 &   &   &  &  &  & $6$ &  \\
				\rowcolor{gray!10}
			 \multirow{-2}{*}{FLIR IR \cite{flirDataset}}&  \multirow{-2}{*}{domain-specific}   & \multirow{-2}{*}{$640\times512$}& \multirow{-2}{*}{$10.3k$}  & \multirow{-2}{*}{$9.3k$}  & \multirow{-2}{*}{15} & ('car', 'motor', 'bus', 'truck',  'scooter', 'other vehicle')    &  \multirow{-2}{*}{76946}\\
			\hline
		\end{tabular}
	}
\end{table*}

For the combination of datasets in our experiments, we choose the following datasets to be included in the training set. From the category of \emph{general} datasets, the MS COCO dataset is included. The basic object detector YOLOv7-X is also trained on MS COCO. From the category of \emph{domain-specific} datasets, we use one dataset from autonomous driving, recorded from ground-level, the FLIR dataset. We split this dataset into sub-sets along the spectral range (infrared (IR) and visual-optical (VIS)). Thus the sub-set of the FLIR dataset gets separate labels for the affinity prediction. From the category of low-level altitude dataset, recorded from a surveillance cam, the DETRAC dataset \cite{Wen_CVIU_2020} is included. The boundary between different altitudes to categorize datasets is not sharp. So, from the category of mid to high-level altitude datasets, we include the VisDrone \cite{Zhu_TPAMI_2021} and UAVDT dataset \cite{Du_ECCV_2018}. The key characteristics of the selected datasets are summarized in Table \ref{tab:selection_datasets}. This includes the considered vehicle child classes that are mapped to the super-category 'vehicle'. 

In addition to aligning the labels, datasets, where the ratio of average object height to image resolution differs strongly from the ratio present in \emph{general} object datasets, are further adapted. These datasets are the \emph{domain-specific} dataset from the categories above ground-level (DETRAC, VisDrone, UAVDT). For these datasets, the original images are sliced into overlapping patches in the range of $600$ to $800$ pixels (see for example Akyon et al. \cite{Akyon_ICIP_2022}). The patches also fit better the image resolution of the basic detector of $640\times640$ pixel. The actual patch size is randomly sampled. We set an overlap ratio of $0.1$. For datasets containing video data, such as DETRAC, we only add every $20$th frame in the training set to prevent including extremely correlated frames. Further, image regions that are not annotated but masking information is provided are colored gray, preventing negative effects during training. Example images from the selected aligned datasets are shown in Figure \ref{fig:examples_dataset}. It should be noted that the number of instances is not completely balanced, but every dataset contains at least over $30k$ instances. Due to the fact that the number of instances per image varies, perfect balancing is also difficult to achieve.
\section{\uppercase{Evaluation}}
\label{sec:evaluation} 
The evaluation is done on unseen datasets to assess the generalization and robustness of the object detector. In the experiments, we use our own dataset captured with the \emph{Fraunhofer} measuring vehicle MODISSA \cite{Borgmann_AO_2021} and the publically available Multi-Spectral Object Detection dataset (MSOD) \cite{Karasawa_ACM_2017}.
\input{figs/fig_Modissa_measument_vehicle}

The MODISSA measuring vehicle is equipped with a range of sensors as well as a visible and infrared panoramic camera setup. For the recording of the test datasets, only the front cameras are used. The visible cameras (VIS) are FLIR Blackfly S BFS-PGE-19S4C with a resolution of $1616\times1240$ pixel, and the infrared cameras (IR) are Device-ALab SmartIR1M0E with a of $1024\times768$ pixel. Figure \ref{fig:modissa} shows the 
MODISSA measurement vehicle with a detailed view of the sensor suite at the front of the vehicle.
\begin{table*}[ht!]
	\caption{Key characteristic of the MODISSA \cite{Borgmann_AO_2021} test datasets used for evaluation.}
	\label{tab:test_datasets} 
	\centering
	\resizebox{1.0\textwidth}{!}{
		\begin{tabular}{|c|c|c|c|c|c|c|}
			\hline
			{\bf dataset} &dataset type & resolution / pixel &\# images  & \# categories &  \# vehicle categories & \# instances  \\
			\hline
			\rowcolor{gray!20}
			  &  &  & & &  $4$ &  \\		
			\rowcolor{gray!20}		
				\multirow{-2}{*}{MODISSA \emph{Vogelsang} (VIS)}  & 	\multirow{-2}{*}{domain-specific} & 	\multirow{-2}{*}{$1616\times1240$} & 	\multirow{-2}{*}{$10k$} & 	\multirow{-2}{*}{$6$} &  ('car', 'motorcycle', 'bus', 'truck') & 	\multirow{-2}{*}{$18214$} \\
				\rowcolor{gray!10}
			&  &  & & &  $4$ &  \\	
			\rowcolor{gray!10}
				\multirow{-2}{*}{MODISSA \emph{Vogelsang} (IR)}  & 	\multirow{-2}{*}{domain-specific} & 	\multirow{-2}{*}{$1024\times768$} & 	\multirow{-2}{*}{$8k$} & 	\multirow{-2}{*}{$6$} &  ('car', 'motorcycle', 'bus', 'truck') & 	\multirow{-2}{*}{$10250$} \\
				\rowcolor{gray!20}
			&  &  & & &  $4$ &  \\	
			\rowcolor{gray!20}
				\multirow{-2}{*}{MODISSA \emph{Realfahrt} (VIS)}  & 	\multirow{-2}{*}{domain-specific} & 	\multirow{-2}{*}{$1616\times1240$} & 	\multirow{-2}{*}{$5.3k$} & 	\multirow{-2}{*}{$6$} &  ('car', 'motorcycle', 'bus', 'truck') & 	\multirow{-2}{*}{$6530$} \\
            \rowcolor{gray!10}
            &  &  & & &  $4$ &  \\	
			\rowcolor{gray!10}
				\multirow{-2}{*}{MODISSA \emph{Realfahrt} (IR)}  & 	\multirow{-2}{*}{domain-specific} & 	\multirow{-2}{*}{$1024\times768$} & 	\multirow{-2}{*}{$5.3k$} & 	\multirow{-2}{*}{$6$} &  ('car', 'motorcycle', 'bus', 'truck') & 	\multirow{-2}{*}{$6530$} \\
			\hline
		\end{tabular}
	}
\end{table*}

The test datasets consist of two different recordings called \emph{Vogelsang} and \emph{Realfahrt}, where we distinguish between the spectral ranges. Thus, four test datasets are separately evaluated. \emph{Vogelsang} captures mainly a residential area with parked vehicles and road traffic. For the \emph{Realfahrt}, only one IR and VIS camera pair is used. The dataset shows further scenes with road traffic and a parking lot. There are more dynamic objects than in the \emph{Vogelsang} dataset. To ensure privacy-preserving (e.g., image recordings of license plates) and complying with corresponding guidelines, we follow the data protection concept of M{\"u}nch et al. \cite{Muench_CVS_2029,Grosselfinger_SPIE_2019}. The annotations contain six classes with four vehicle classes ('car', 'motorcycle', 'bus', 'truck'). Similar to the training data, these sub-categories are mapped to one 'vehicle' class. All key characteristics of the MODISSA test datasets are summarized in Table \ref{tab:test_datasets}.
\input{figs/fig_example_detection_modissa}
\begin{table*}[h!]
	\caption{Key characteristic of the MSOD \cite{Karasawa_ACM_2017} test datasets used for evaluation.}
	\label{tab:test_datasets_MSOD} 
	\centering
	\resizebox{1.0\textwidth}{!}{
		\begin{tabular}{|c|c|c|c|c|c|c|}
			\hline
			{\bf dataset} & dataset type & resolution / pixel &\# images  & \# categories &  \# vehicle categories & \# instances  \\
			\hline
			\rowcolor{gray!20}
			&  &  & & &  $1$ &  \\	
			\rowcolor{gray!20}		
				\multirow{-2}{*}{MSOD (VIS) \cite{Karasawa_ACM_2017}} & 	\multirow{-2}{*}{domain-specific} & 	\multirow{-2}{*}{$640\times480$} &	\multirow{-2}{*}{$7.5k$} & 	\multirow{-2}{*}{$9$} & ('car') & 	\multirow{-2}{*}{$7426$} \\
				\rowcolor{gray!10}
			&  &  & & &  $1$ &  \\	
			\rowcolor{gray!10}
				\multirow{-2}{*}{MSOD (NIR) \cite{Karasawa_ACM_2017}} & 	\multirow{-2}{*}{domain-specific} & 	\multirow{-2}{*}{$320\times256$} & 	\multirow{-2}{*}{$7.5k$} & 	\multirow{-2}{*}{$9$} & ('car') & 	\multirow{-2}{*}{$5209$} \\
			\rowcolor{gray!20}
			&  &  & & &  $1$ &  \\	
			\rowcolor{gray!20}
				\multirow{-2}{*}{MSOD (MIR) \cite{Karasawa_ACM_2017}} & 	\multirow{-2}{*}{domain-specific} & 	\multirow{-2}{*}{$320\times256$} & 	\multirow{-2}{*}{$7.5k$} & 	\multirow{-2}{*}{$9$} &  ('car') & 	\multirow{-2}{*}{$4472$} \\
			\rowcolor{gray!10}
			&  &  & & &  $4$ &  \\	
			\rowcolor{gray!10}
			\multirow{-2}{*}{MSOD (FIR) \cite{Karasawa_ACM_2017}}  & 	\multirow{-2}{*}{domain-specific} & 	\multirow{-2}{*}{$640\times480$} & 	\multirow{-2}{*}{$7.5k$} & 	\multirow{-2}{*}{$9$} &  ('car') & 	\multirow{-2}{*}{$5042$} \\
			\hline
		\end{tabular}
	}
\end{table*}

The selected public MSOD dataset \cite{Karasawa_ACM_2017} is a \emph{domain-specific}  dataset for autonomous driving that consists of multi-spectral (VIS, NIR, MIR, and FIR) images. Similarity to the MODISSA dataset, the different spectral ranges are separately evaluated.  The nine original ground truth class labels include only one class ('car') mapped to 'vehicle'. The images show traffic scene in an university environment at daytime and nighttime. All key characteristics of the MSOD test datasets are summarized in Table \ref{tab:test_datasets_MSOD}.

To quantify the results, we use the \emph{mean average precision} (mAP) object detector metrics. In particular, mAP@.5 \cite{everingham2010pascal} and mAP@.5:.95 \cite{Lin_ECCV_2014} are used. While mAP@.5 is the mAP for an IoU threshold of at least fifty percent, the mAP@.5:.95 is the average across ten IoU thresholds, hence more strict.   

After aligning the label sets as described in section \ref{sec:dataset_alignment}, we first train a 'vehicle' detector with the selected six datasets (MS COCO, DETRAC, VisDrone, UAVDT, FLIR VIS, FLIR IR) for the experiments. Besides calculating the mAP values, the maximum score of the affinity prediction is used to estimate the contributing dataset of a true positive (TP) detection. The distribution of assigned training datasets of truly detected vehicles is calculated over the evaluation dataset. It is used to split the training into the sets of the two highest assigned datasets and the remaining datasets. After training on the split sets, the performance between detectors is compared. The quantitative results of these experiments for the MODISSA datasets are shown in Table \ref{tab:results_expermints}, and some exemplary qualitative results are visualized in Figure \ref{fig:results}. Detections assigned to MS COCO are highlighted in red  \cbox{Fraunhoferred}. The assigned FLIR IR detections are shown in aqua \cbox{Fraunhoferaqua} and detections assigned to FLIR VIS are shown in lime \cbox{Fraunhoferlime}. 
\begin{table*}[ht!] 
\caption{Comparison of reference detectors trained on different combinations of aligned datasets for the MODISSA datasets. In addition, the percentage of assigned datasets for true positive detection based on the dataset affinity score is depicted.}
\resizebox{\textwidth}{!}{
	\begin{tabular}{| c | c c | c c c c c c | c c |}
      	\hline
     	\multicolumn{1}{|c|}{test dataset} & \multicolumn{2}{c|}{ spectral range} &\multicolumn{6}{c|}{ training datasets + percentage of assigned true positive detection}  & \multicolumn{2}{c|}{mAP $\uparrow$ }  \\
		\multicolumn{1}{|c|}{} & {VIS} & {IR} & {MS COCO} & {DETRAC} & {UAVDT} & {VisDrone} & {FLIR VIS} & {FLIR IR} &    mAP@.5 & mAP@.5-.95   \\
		\hline
		\rowcolor{gray!20}
		MODISSA \emph{Vogelsang}            & \color{Fraunhofergreen}{\cmark} & \xmark  & \color{Fraunhofergreen}{\cmark (45.6\%)}  & \color{Fraunhofergreen}{\cmark (0.1\%)} & \color{Fraunhofergreen}{\cmark ($\approx$} 0\%)   & \color{Fraunhofergreen}{\cmark ($\approx$ 0\%)} & \textbf{\color{Fraunhofergreen}{\cmark (54.3\%)}} & \color{Fraunhofergreen}{\cmark ($\approx$ 0\%)} & 0.921 &  0.802    \\
		\rowcolor{gray!10}
		MODISSA \emph{Vogelsang}               & \color{Fraunhofergreen}{\cmark} & \xmark & \color{Fraunhofergreen}{\cmark}(44.7\%)  & \xmark (-\%) & \xmark (-\%)&  \xmark (-\%)& \textbf{\color{Fraunhofergreen}{\cmark}(55.3\%)} &  \xmark (-\%) &  0.913   &    0.785     \\
		\rowcolor{gray!20}
	   	MODISSA \emph{Vogelsang}      & \color{Fraunhofergreen}{\cmark} & \xmark & \xmark (-\%)  & \textbf{\color{Fraunhofergreen}{\cmark (99.7\%)}}  & \color{Fraunhofergreen}{\cmark (0.1\%)} & \color{Fraunhofergreen}{\cmark (0.1\%)} & \xmark (-\%) & \color{Fraunhofergreen}{\cmark (0.1\%)} &  0.662   &  0.482     \\	
	   	 \hline		
		\rowcolor{gray!10}
  	 	MODISSA \emph{Vogelsang}               & \xmark & \color{Fraunhofergreen}{\cmark}  & \color{Fraunhofergreen}{\cmark (1.2\%)}  & \color{Fraunhofergreen}{\cmark (0.1\%)} & \color{Fraunhofergreen}{\cmark (0.1\%)} & \color{Fraunhofergreen}{\cmark (0.1\%)} & \color{Fraunhofergreen}{\cmark (0.1\%)} & \textbf{\color{Fraunhofergreen}{\cmark (98.4\%)}}&  0.914    &    0.780     \\
		\rowcolor{gray!20}
       	MODISSA \emph{Vogelsang}               & \xmark & \color{Fraunhofergreen}{\cmark} & \color{Fraunhofergreen}{\cmark (0.6\%)}  &  \xmark (-\%) &  \xmark (-\%) &  \xmark (-\%) &   \xmark (-\%)& \textbf{\color{Fraunhofergreen}{\cmark (99.4\%)}} & 0.911   &    0.781     \\
		\rowcolor{gray!10}
       	MODISSA \emph{Vogelsang}               & \xmark & \color{Fraunhofergreen}{\cmark} & \xmark (-\%)  & \textbf{\color{Fraunhofergreen}{\cmark (99.1 \%)}} & \color{Fraunhofergreen}{\cmark (0.3\%)}  & \color{Fraunhofergreen}{\cmark (0.3\%)}&  \color{Fraunhofergreen}{\cmark (0.1\%)}&  \xmark (-\%)&  0.528    &   0.396     \\
        \hline
      	\rowcolor{gray!20}
      	MODISSA \emph{Realfahrt}               & \color{Fraunhofergreen}{\cmark} & \xmark  & \color{Fraunhofergreen}{\cmark (29.5\%)}  & \color{Fraunhofergreen}{\cmark ($\approx$ 0\%)} & \color{Fraunhofergreen}{\cmark ($\approx$ 0\%)} & \color{Fraunhofergreen}{\cmark ($\approx$ 0\%)}& \textbf{\color{Fraunhofergreen}{\cmark (70.5\%)}} & \color{Fraunhofergreen}{\cmark ($\approx$ 0\%)} & 0.822 & 0.617     \\
        \rowcolor{gray!10}
      	MODISSA \emph{Realfahrt}               & \color{Fraunhofergreen}{\cmark} & \xmark & \color{Fraunhofergreen}{\cmark (30.8\%)}   & \xmark (-\%)  & \xmark (-\%) & \xmark (-\%) & \textbf{\color{Fraunhofergreen}{\cmark ( 69.2\%)}} & \xmark (-\%) & 0.780   &    0.561     \\
        \rowcolor{gray!20}
      	MODISSA \emph{Realfahrt}       & \color{Fraunhofergreen}{\cmark} & \xmark &  \xmark (-\%) & \textbf{\color{Fraunhofergreen}{\cmark (99.4 \%)}}  & \color{Fraunhofergreen}{\cmark (0.4 \%)}  & \color{Fraunhofergreen}{\cmark (0.2 \%)} & \xmark (-\%) & \color{Fraunhofergreen}{\cmark ($\approx$ 0 \%)}   & 0.716  &   0.416  \\			
        \hline
        \rowcolor{gray!10}
        MODISSA \emph{Realfahrt}  & \xmark & \color{Fraunhofergreen}{\cmark}  &  \color{Fraunhofergreen}{\cmark (1.2 \%)}  & \color{Fraunhofergreen}{\cmark ( 0.1\%)}  & \color{Fraunhofergreen}{\cmark (0.1 \%)}  &  \color{Fraunhofergreen}{\cmark ( 0.1\%)} & \color{Fraunhofergreen}{\cmark ( 3.9\%)} & \textbf{\color{Fraunhofergreen}{\cmark ( 94.6\%)}}  & 0.674    &   0.400     \\
        \rowcolor{gray!20}
        MODISSA \emph{Realfahrt} & \xmark & \color{Fraunhofergreen}{\cmark} & \color{Fraunhofergreen}{\cmark (1.9 \%)}  & \xmark (-\%) & \xmark (-\%) & \xmark (-\%)&  \xmark (-\%) &  \textbf{\color{Fraunhofergreen}{\cmark ( 98,1\%)}} &  0.673    & 0.400 \\
        \rowcolor{gray!10}
        MODISSA \emph{Realfahrt}       & \xmark & \color{Fraunhofergreen}{\cmark} &  \xmark (-\%) & \textbf{\color{Fraunhofergreen}{\cmark (99.2 \%)}} & \color{Fraunhofergreen}{\cmark ( 0.3\%)} & \color{Fraunhofergreen}{\cmark (0.3 \%)}& \color{Fraunhofergreen}{\cmark ( 0.2\%)} &  \xmark (-\%) & 0.449  &     0.238    \\
	   	\hline
	\end{tabular} 
	}	
\label{tab:results_expermints}
\end{table*} 

The results show that the best performance could be achieved by using all the complete aligned training data. This applies to all MODISSA test datasets. Thus, increasing the variation and number of training samples helped to generalize to these datasets. Hence, the concept of merging datasets has also here proven to learn more robust and general models. When looking at the distribution of the assigned dataset by the affinity scores, one can see that for the VIS \emph{Vogelsang} and \emph{Realfahrt} mainly the MS COCO and FLIR VIS datasets are classified as origin dataset. For the IR \emph{Vogelsang} and \emph{Realfahrt}, almost all true positive detections are classified as originating from the FLIR IR dataset. Since the test dataset is in the application of autonomous driving captured from ground-level, this might not be surprising. Mainly because only one IR dataset is in the training set. However, this can be seen as some sanity check that the proposed idea of adding the dataset affinity prediction as additional inference enables useful feedback over the training set.
\begin{table*}[h!] 
\caption{Comparison of reference detectors trained on different combinations of aligned datasets for the MSOD datasets. In addition, the percentage of assigned datasets for true positive detection based on the dataset affinity score is depicted.}
\resizebox{\textwidth}{!}{
	\begin{tabular}{| c | c c c c| c c c c c c | c c |}
      	\hline
     	\multicolumn{1}{|c|}{test dataset} & \multicolumn{4}{c|}{ spectral range} &\multicolumn{6}{c|}{ training datasets + percentage of assigned true positive detection}  & \multicolumn{2}{c|}{mAP $\uparrow$ }  \\
		\multicolumn{1}{|c|}{} & {VIS} & {NIR} & {MIR} & {FIR} & {MS COCO} & {DETRAC} & {UAVDT} & {VisDrone} & {FLIR VIS} & {FLIR IR} &    mAP@.5 & mAP@.5-.95   \\
		\hline
		\rowcolor{gray!20}
		MSOD   & \color{Fraunhofergreen}{\cmark} & \xmark & \xmark & \xmark & \textbf{\color{Fraunhofergreen}{ \cmark (86.3\%)}}  & \color{Fraunhofergreen}{\cmark (0.1\%)} & \color{Fraunhofergreen}{\cmark ($\approx$} 0\%)   & \color{Fraunhofergreen}{\cmark (4.0\%)} & \color{Fraunhofergreen}{\cmark (9.6\%)} & \color{Fraunhofergreen}{\cmark ($\approx$ 0\%)} & 0.490 &  0.293    \\
		\rowcolor{gray!10}
		MSOD     & \color{Fraunhofergreen}{\cmark} & \xmark & \xmark & \xmark & \textbf{\color{Fraunhofergreen}{\cmark}(88.2\%)}  & \xmark (-\%) & \xmark (-\%)&  \xmark (-\%)& \color{Fraunhofergreen}{\cmark}(11.8\%) &  \xmark (-\%) &  0.487   &  0.280     \\
		\rowcolor{gray!20}
	   	MSOD  & \color{Fraunhofergreen}{\cmark} & \xmark  & \xmark  & \xmark & \xmark (-\%)  & \textbf{\color{Fraunhofergreen}{\cmark (99.3\%)}}  & \color{Fraunhofergreen}{\cmark (0.1\%)} & \color{Fraunhofergreen}{\cmark (0.5\%)} & \xmark (-\%) & \color{Fraunhofergreen}{\cmark (0.1\%)} &  0.314   &    0.177     \\	
	   	 \hline		
		\rowcolor{gray!10}
  	 	MSOD   & \xmark & \color{Fraunhofergreen}{\cmark}              & \xmark & \xmark & \textbf{\color{Fraunhofergreen}{\cmark (49.1\%)}}  & \color{Fraunhofergreen}{\cmark (0.8\%)} & \color{Fraunhofergreen}{\cmark (0.1\%)} & \color{Fraunhofergreen}{\cmark (0.1\%)} & \color{Fraunhofergreen}{\cmark (43.4\%)} & \color{Fraunhofergreen}{\cmark (6.5\%)}&   0.458    &   0.260     \\
		\rowcolor{gray!20}
       	MSOD   & \xmark & \color{Fraunhofergreen}{\cmark}            & \xmark & \xmark & \color{Fraunhofergreen}{\cmark (38.8\%)}  &  \xmark (-\%) &  \xmark (-\%) &  \xmark (-\%) &   \textbf{\color{Fraunhofergreen}{\cmark (61.2\%)}} &  \xmark (-\%) & 0.429   &   0.238     \\
		\rowcolor{gray!10}
       	MSOD  & \xmark & \color{Fraunhofergreen}{\cmark}                & \xmark & \xmark & \xmark (-\%)  & \textbf{\color{Fraunhofergreen}{\cmark (99.6 \%)}} & \color{Fraunhofergreen}{\cmark (0.2\%)}  & \color{Fraunhofergreen}{\cmark (0.1\%)}& \xmark (-\%) &  \color{Fraunhofergreen}{\cmark (0.1\%)} &  0.284   &    0.154     \\
        \hline
      	\rowcolor{gray!20}
       	MSOD  & \xmark & \xmark & \color{Fraunhofergreen}{\cmark} & \xmark  & \color{Fraunhofergreen}{\cmark (1.8\%)}  & \color{Fraunhofergreen}{\cmark ($\approx$ 0\%)} & \color{Fraunhofergreen}{\cmark ($\approx$ 0\%)} & \color{Fraunhofergreen}{\cmark ($0.8$\%)} & \color{Fraunhofergreen}{\cmark ($\approx$ 0\%)} & \textbf{\color{Fraunhofergreen}{\cmark (97.4\%)}} & 0.496   &   0.316    \\
        \rowcolor{gray!10}
      	MSOD   & \xmark & \xmark  & \color{Fraunhofergreen}{\cmark} & \xmark & \color{Fraunhofergreen}{\cmark (7.2\%)}   & \xmark (-\%)  & \xmark (-\%) & \xmark (-\%) & \xmark (-\%) &  \textbf{\color{Fraunhofergreen}{\cmark ( 92.8\%)}} & 0.489   &    0.306   \\
        \rowcolor{gray!20}
      	MSOD & \xmark & \xmark  & \color{Fraunhofergreen}{\cmark} & \xmark &  \xmark (-\%) & \textbf{\color{Fraunhofergreen}{\cmark (98.8 \%)}}  & \color{Fraunhofergreen}{\cmark (0.4 \%)}  & \color{Fraunhofergreen}{\cmark (0.8 \%)} & \color{Fraunhofergreen}{\cmark ($\approx$ 0 \%)} & \xmark (-\%)   & 0.239  &    0.141  \\			
        \hline
        \rowcolor{gray!10}
        MSOD  & \xmark & \xmark  & \xmark & \color{Fraunhofergreen}{\cmark}  &  \color{Fraunhofergreen}{\cmark (0.3 \%)}  & \color{Fraunhofergreen}{\cmark ($\approx$ 0\%)}  & \color{Fraunhofergreen}{\cmark ($\approx$ 0\%)} &  \color{Fraunhofergreen}{\cmark ($\approx$ 0\%)} & \color{Fraunhofergreen}{\cmark ($\approx$ 0\%)} & \textbf{\color{Fraunhofergreen}{\cmark ( 99.7\%)}}  & 0.505  &     0.293  \\
        \rowcolor{gray!20}
        MSOD & \xmark & \xmark  & \xmark & \color{Fraunhofergreen}{\cmark} & \color{Fraunhofergreen}{\cmark (0.2 \%)}  & \xmark (-\%) & \xmark (-\%) & \xmark (-\%)&  \xmark (-\%) &  \textbf{\color{Fraunhofergreen}{\cmark ( 99,8\%)}} &  0.520  &     0.302 \\
        \rowcolor{gray!10}
        MSOD     & \xmark & \xmark    & \xmark & \color{Fraunhofergreen}{\cmark} &  \xmark (-\%) & \textbf{\color{Fraunhofergreen}{\cmark (93.2 \%)}} & \color{Fraunhofergreen}{\cmark ( 3.8\%)} & \color{Fraunhofergreen}{\cmark (2.5 \%)}& \color{Fraunhofergreen}{\cmark ( 0.5\%)} &  \xmark (-\%) & 0.153   &   0.0825   \\
	   	\hline
	\end{tabular} 
	}	
\label{tab:results_expermints_MSOD}
\end{table*} 
\input{figs/fig_results_MSOD} 

Moreover, when we look at the results achieved using only the data from the datasets with the highest percentage of assigned dataset affinity, it becomes visible that the drop in performance is relatively low compared to the full set. In contrast, using the remaining datasets led to a drastic performance drop despite the total number of training images and instances being higher (see Table \ref{tab:selection_datasets}). The fact that from the remaining datasets, DETRAC is then the most assigned dataset also corresponds to the intuition that the low-level altitude dataset is closest to the test domain. 
\input{figs/fig_results_others}

This also applies to the results of the experiments for the MSOD datasets, shown in Table \ref{tab:results_expermints_MSOD}, and some quantitative results are shown in Figure \ref{fig:results_MSOD}. The colors of the bounding boxes encode the assigned dataset. Red \cbox{Fraunhoferred} corresponds to MS COCO, lime \cbox{Fraunhoferlime} to FLIR VIS, and aqua \cbox{Fraunhoferaqua} to FLIR IR. The overall tendency complies with the result of the MODISSA datasets, although there are minor differences. The full training pool achieves the best performance for almost all MSOD test datasets. The only exception is the FIR data, where using only the main supporting datasets achieved even better results. What can be seen from these results is the shift along the spectral range what datasets are responsible for the detections. Whereas for the VIS data, the detector mainly assigns MS COCO and FLIR VIS. The lower mAP values for the MSOD datasets can be explained with the lower image resolution and correspondingly lower object sizes in the image. The shift towards FLIR IR can be seen when considering images corresponding to higher wavelength spectra. Interestingly, the detector still relies on MS COCO and FLIR VIS in the NIR data. A minor difference is that MS COCO is the dataset with the estimated strongest support for the VIS data. Nonetheless, also these results show the proposed dataset affinity score can be used to automatically select samples from a heterogeneous pool of vehicle datasets. Besides, the model is trained on a significantly sparser set of training samples, there is almost no performance decrease and even a counterproductive training data combination could be identified. Since the selected dataset pool spans across different categories, adding a dataset from often assigned categories is a way to optimize an object detector on a specific application.  

The benefit of the ante-hoc detection explanation provided by the affinity prediction can be seen without a quantitative evaluation. By applying the adapted detector to desired target domain data, it is possible to directly get insights how to possible extend the training set and if samples from the target domain can already be explained by included datasets. Figure \ref{fig:results_other} shows some exemplary detections on datasets outside the training domain to depict this effect. Images are taken from the DroneVehicle dataset \cite{sun2020drone}, the 
KAIST Multispectral Pedestrian Dataset (KAIST MPD) \cite{Hwang_CVPR_2015}, the Dense Depth for Automated Driving dataset (DDAD) \cite{Guizilini_CVPR_2020}, the REalistic and Diverse Scenes dataset (REDS) \cite{Nah_CVPRW_2019}, and Aerial Multi-Vehicle Detection Dataset (AMVD). The colors of the bounding boxes encode the assigned dataset. Red \cbox{Fraunhoferred}, blue \cbox{Fraunhofersteelblue}, lime  \cbox{Fraunhoferlime}, and aqua \cbox{Fraunhoferaqua} correspond to respectively MS COCO, VisDrone, FLIR VIS, FLIR IR. For the examples from datasets captured from a high altitude, the VisDrone dataset is assigned. Although the results, together with the chosen set of aligned datasets and the test dataset, follow an intuition of what dataset combination should work, the proposed affinity score can help to find dataset bias and outliers in the data and offers an additional tool to assess the training data.

\section{Conclusion}
\label{sec:conclusion}
In this paper, we proposed to add an additional inference head to an object detection pipeline for predicting the training data affinity. Since current detectors have inherent different heads for separated inference tasks, this extension can be applied to most current detectors. By merging existing datasets to learn a more robust model, we first aligned several datasets toward this end. Then, we evaluated an exemplary detector and used the affinity score to assess the contribution of specific datasets on individual detections. We demonstrated the efficacy of the dataset affinity prediction by achieving comparable results with significantly fewer training samples by focusing on datasets with more substantial support, as indicated by the affinity scores. Moreover, the proposed dataset affinity prediction offers some kind of ante-hoc detection explanation during inference and helps to assess the effectiveness of pooling datasets. 

%
%
%
%
%

\bibliographystyle{splncs04}
\bibliography{Becker_ROBOVIS_2024}

%
%
%
%
\end{document}